\documentclass{article}
\usepackage{spconf,amsmath,graphicx}
\usepackage{enumitem}
\usepackage{url}
\usepackage[bottom]{footmisc}
\usepackage[caption=false,font=footnotesize]{subfig}
\usepackage{amsmath}
\usepackage{amssymb}
\usepackage{amsfonts}
\usepackage{breakcites}
 \usepackage{tabularx}
 \usepackage{multirow}
 
\usepackage{graphicx}
\usepackage{amsmath}
\usepackage{amssymb}
\usepackage{amsthm,bm}
\usepackage{float}
\usepackage{tikz}
\usepackage{xcolor}
\usepackage{optidef}
\usepackage[normalem]{ulem}
\usepackage{algorithm}
\usepackage{algpseudocode}
 \usepackage{url}
 \usepackage[outdir=./]{epstopdf}

\newcommand{\bx}{\mathbf{x}}

\newcommand{\bv}{\mathbf{v}}

\newcommand{\bh}{\mathbf{h}}

\newcommand{\bY}{\mathbf{Y}}

\newcommand{\bV}{\mathbf{V}}

\newcommand{\bs}{\mathbf{s}}

\newcommand{\bS}{\mathbf{S}}

\newcommand{\bH}{\mathbf{H}}

\theoremstyle{definition}

\graphicspath{{./Figures/}}

\usepackage{graphicx}

\title{A Bayesian Perspective for Determinant Minimization Based Robust Structured Matrix Factorization}
\twoauthors{Gokcan Tatli}{ECE Department, \\University of Wisconsin-Madison, WI USA}{Alper T. Erdogan}{ KUIS AI Center, and EEE Department,\\ Koc University, Istanbul, Turkey }
\begin{document}
%\ninept
%
\maketitle
\begin{abstract}
We introduce a Bayesian perspective for the structured matrix factorization problem. The proposed framework provides a probabilistic interpretation for existing geometric methods based on determinant minimization. We model input data vectors as linear transformations of latent vectors drawn from a distribution uniform over a particular domain reflecting structural assumptions, such as the probability simplex in Nonnegative Matrix Factorization and polytopes in Polytopic Matrix Factorization. We represent the rows of the linear transformation matrix as vectors generated independently from a normal distribution whose covariance matrix is inverse Wishart distributed. We show that the corresponding maximum a posteriori estimation problem boils down to the robust determinant minimization approach for structured matrix factorization, providing insights about parameter selections and potential algorithmic extensions.
\end{abstract}
%

% ,-.          .                ,       ,     .             .         .             
%(   `         |   o           '|   o   |     |             |         |   o         
% `-.  ,-. ,-. |-  . ,-. ;-.    |       | ;-. |-  ;-. ,-. ,-| . . ,-. |-  . ,-. ;-. 
%.   ) |-' |   |   | | | | |    |   o   | | | |   |   | | | | | | |   |   | | | | | 
% `-'  `-' `-' `-' ' `-' ' '    '       ' ' ' `-' '   `-' `-' `-` `-' `-' ' `-' ' ' 

\section{Introduction}
\label{sec:intro}

Structured matrix factorization (SMF) is a significant area of interest in signal processing, machine learning, and data science with a diverse set of application areas \cite{paatero1994positive,koren2009matrix,xu2003document,yokoya2011coupled,cichocki2009nonnegative, tatli:21tsp}. SMF methods model the input data matrix $\bY$ as the product of some unknown factor matrices $\bH$ and $\bS$ that contain useful latent information. 

Obtaining original factors from input data is a significant challenge that requires exploiting particular assumptions. For example, we typically assume that $\bH$ is full column rank and that the columns of $\bS$ are located in a particular domain. The choice of this domain determines the type of factorization, e.g., probability simplex in simplex-structured matrix factorization (SSMF) or nonnegative matrix factorization (NMF)\cite{fu2019nonnegative},  and polytopes in polytopic matrix factorization (PMF)\cite{tatli:21tsp}. Assuming that the columns of $\bS$ are "sufficiently scattered" \cite{fu2015blind,fu2018identifiability,tatli:21tsp} in their domain,  the minimization of $\det(\bH^T\bH)$ emerges as an effective optimization strategy to extract the original factors. This strategy is mainly motivated by convex geometry, which can also be backed by probabilistic models built on this geometric foundation, particularly for the probability simplex setting \cite{fu2019nonnegative, Wu:2022}. Although the determinant optimization criterion was successfully extended to polytopic domains \cite{tatli:21tsp, tatli:21icassp}, its volume-based justification and its probabilistic extensions are not directly applicable. Therefore, our article aims to provide an alternative framework in which the determinant minimization for SMF (both SSMF and PMF) can be posed as a maximum a posteriori estimation.

\subsection{Related Works and Contributions}
There already exist some frameworks for probabilistic interpretations or extensions of the structured matrix factorization based on volume/determinant minimization, for which we provide more details in Section \ref{sec:review}. An earlier reference by Arngren et al. \cite{arngren2009bayesian} proposes a Bayesian framework for the determinant minimization-based NMF. This approach makes direct use of a prior probability density function (pdf) for the factor $\bH$, in the form of a maximum entropy density with a constraint on the determinant of $\bH^T\bH$. The proposed distribution reflects the volume-based prior for the corresponding simplex geometry. More recently, Wu et al. proposed a maximum likelihood framework for simplex component analysis \cite{Wu:2022}, in which determinant minimization is again connected to the volume of the simplex defined by the data model with nonnegativity constraint and emerges as the approximation of the likelihood function in the noisy case.

In this article, we propose a novel Bayesian framework in which the rows of $\bH$ are assumed to be independently drawn from a multivariate normal distribution with an unknown covariance matrix. We also assume an inverse Wishart distribution for the covariance matrix, the conjugate prior for the Gaussian observation model. Therefore, for the proposed framework,
\begin{itemize}
\item dissimilar to the Bayesian framework in \cite{arngren2009bayesian}, we do not use a volume-prior-based density for $\bH$,
\item unlike the ML approach in \cite{Wu:2022}, the formulation of determinant minimization is a direct consequence, not an approximation of the objective of the original probabilistic framework in the noisy scenario,
\item different from both \cite{arngren2009bayesian} and \cite{Wu:2022}, the determinant minimization is obtained for more general source domains, including polytopes \cite{tatli:21tsp}, for which there is no explicit formula for the volume in terms of the determinant, unlike the simplex domains.
\end{itemize}

% ,-.          .               ,-.        ,-.                      
%(   `         |   o              )   o   |  )         o           
% `-.  ,-. ,-. |-  . ,-. ;-.     /        |-<  ,-. . , . ,-. , , , 
%.   ) |-' |   |   | | | | |    /     o   |  \ |-' |/  | |-' |/|/  
% `-'  `-' `-' `-' ' `-' ' '   '--'       '  ' `-' '   ' `-' ' '   

\section{A Review of determinant minimization Based Matrix Factorization Frameworks}
\label{sec:review}

\subsection{Deterministic Structured Matrix Factorization}
In the conventional deterministic structured matrix factorization problem, we assume a generative setting in which the input matrix $\bY \in \mathbb{R}^{M\times N}$ is given by
\begin{eqnarray}
 \bY=\bH_g\bS_g, \label{eq:SMFgenba}
\end{eqnarray}
where
%\begin{itemize}
%\item
$\bH_g\in \mathbb{R}^{M \times r}$ is the ground truth of the left-factor matrix, which is assumed to be of full-column-rank; and %In the Blind Source Separation (BSS) application, this would correspond to the mixing matrix..
%\item
$\bS_g\in \mathbb{R}^{r \times N}$ is the ground truth right-factor matrix that contains latent vectors, where we assume $r\le \min(M,N)$ and
\begin{eqnarray}
 {\bS_g}_{:,j} \in \mathcal{D}, \; j= 1, \ldots, N, \label{eq:SMFgen2ba}
\end{eqnarray}
where $\mathcal{D}$ is the domain assumed for the columns of $\bS_g$, such as the probability simplex, 
%\begin{eqnarray}
$\Delta_r=\{\bx \hspace{0.05in}\vert\hspace{0.05in} \mathbf{1}^T\bx=1, \bx\ge 0, \bx \in \mathbb{R}^r\}$,
%\end{eqnarray}
in the NMF and SSMF frameworks and polytope,
%\begin{eqnarray}
$\mathcal{P}=\text{conv}(\{\bv_1, \bv_2, \ldots, \bv_m\})$,
%\end{eqnarray}
in the PMF framework. 
%\end{itemize}

%The determinant maximization criterion, which is defined in terms of the optimization problem,
%\begin{maxi!}[l]<b>
%{\bH\in \mathbb{R}^{M \times r},\bS\in \mathbb{R}^{r\times N}}
%{\det(\bS\bS^T)\label{eq:detmaxobjective}}{\label{eq:detmaxoptimization}}{}
%\addConstraint{\bY=\bH \bS}{\label{eq:detmaxconstr1}}{}
%\addConstraint{\bS_{:,j} \in \mathcal{D}}{,\quad}{j= 1, \ldots, N,\label{eq:detmaxconstr2}}{ }
%\end{maxi!}
The determinant minimization criterion has been proposed to extract the estimates of the original factors from the input data matrix $\bY$, in the NMF, SSMF \cite{chan2011simplex,fu2018identifiability,fu2019nonnegative} and PMF \cite{tatli:21tsp} frameworks, through the optimization setting,
\begin{mini!}[l]<b>
{\bH\in \mathbb{R}^{M \times r},\bS\in \mathbb{R}^{r\times N}}
{\det(\bH^T\bH)\label{eq:detminobjective}}{\label{eq:detminoptimization}}{}
\addConstraint{\bY=\bH \bS}{\label{eq:detminconstr1}}{}
\addConstraint{\bS_{:,j} \in \mathcal{D}}{,\quad}{j= 1, \ldots, N,\label{eq:detminconstr2}}{ }
\end{mini!} 
which is motivated by finding the simplex with the minimum volume, the objective in (\ref{eq:detminobjective}) that contains the input vectors. 
Under the assumption that the columns of $\bS_g$ in (\ref{eq:SMFgen2ba}) are sufficiently scattered \cite{fu2018identifiability, fu2019nonnegative,tatli:21tsp} in $\mathcal{D}$, the global optima of (\ref{eq:detminoptimization}) recover the original factors $(\bH_g,\bS_g)$ up to some uncertainties in the permutations and/or signs of the columns (rows) of $\bS_g$ ($\bH_g$).\looseness=-1

\subsection{Probabilistic Models for Determinant Minimization Criterion}
The determinant minimization criterion in (\ref{eq:detminoptimization}), introduced for the deterministic setting in (\ref{eq:SMFgenba})-(\ref{eq:SMFgen2ba}), is attractive due to its geometric foundations, as well as finite sample identifiability guarantees that it provides under {\it sufficiently scattering} conditions \cite{fu2018identifiability,fu2019nonnegative,tatli:21tsp,tatli:21icassp}. However, stochastic interpretations of this criterion would be useful in terms of employing available probabilistic analysis tools and factoring potential noise effects in input data.

\subsubsection{A Bayesian Approach for NMF}
For NMF, Arngren et al. provide a Bayesian interpretation for determinant minimization via the use of the following direct prior pdf assumption on the left factor $\bH$ \cite{arngren2009bayesian}:
\begin{eqnarray}
f_{\bH}(\bH)=\left\{\begin{array}{cc}  e^{-\gamma \det(\bH^T\bH)} & \bH\ge 0, \\ 0  & \text{otherwise,} \end{array}\right.
\end{eqnarray}
which is motivated as the volume prior assumption.
The same article also proposes the use of the additive Gaussian noise model for observations, that is, (\ref{eq:SMFgenba}) is replaced with
\begin{eqnarray}
 \bY=\bH_g\bS_g+\bV_g, \label{eq:SMFgenbaNoisy}
\end{eqnarray}
where the elements of the noise sample $\bV_g$ are drawn independently from the distribution $\mathcal{N}(0, \sigma_v^2)$. In the same model, the latent vectors $\{{\bS_g}_{:,j}, j=1, \ldots N\}$ are independently drawn from the uniform distribution 
\begin{eqnarray*}
f_{U}(\bs)=\frac{1}{\text{vol}(\Delta_r)}{1}_{{\Delta}_r}(\bs), \hspace{0.03in}{1}_{{\Delta}_r}(\bs)=\left\{\begin{array}{cc} 1 & \bs \in \Delta_r, \\ 0  & \text{otherwise.} \end{array}\right.%, i.e., the indicator function. 
\end{eqnarray*}
%\end{eqnarray}

\subsubsection{A Maximum Likelihood Approach for Simplex Structured Matrix Factorization}
\looseness=-1 Recently, Wu et al. proposed a maximum likelihood framework, called PRISM (probabilistic simplex component analysis), for simplex-structured matrix factorization \cite{Wu:2022}. This framework is also based on the noisy input data model in (\ref{eq:SMFgenbaNoisy}). However, $\bH_g$ is modeled as an unknown deterministic parameter, rather than a random quantity in a Bayesian setting. For the columns of the matrix $\bS_g$, we assume Drichlet distribution, of which particular case is  uniform distribution.
%\begin{eqnarray}
%f_{D}(\bs;\mathbf{\alpha})= \frac{1}{B(\mathbf{\alpha})}\left(1-\sum_{i=1}^{r-1}s_i\right)^{\alpha_r-1}\prod_{i=1}^{r-1}s_i^{\alpha_i-1}{1}_{\bar{\Delta}_r}(\bs), \label{eq:drichdist}
%\end{eqnarray}
%where $\bar{\Delta}_r$ is the open unit simplex defined as $\bar{\Delta}_r=\{\bs \in \mathbb{R}^r \hspace{0.05in} \vert \hspace{0.05in} \mathbf{1}^T\bs=1\}$. 
For the no-noise case, the likelihood maximization is shown to boil down to minimum volume data-enclosing simplex problem:
\begin{mini}
{\bH\in \mathbb{R}^{M \times r}}
{\text{vol}(\bH)}{\label{eq:minvoloptimization}}{}
\addConstraint{\bY_{:,i} \in \overline{\text{conv}}(\bH)}{}{}
\end{mini}
where $\overline{\text{conv}}(\bH)=\{ \bH\bs \hspace{0.05in}\vert\hspace{0.05in} \bs\in\mathbb{R}_{++}^r, \mathbf{1}^T\bs=1\}$, $\text{vol}(\bH)=\frac{\det({\bar{\bH}}^T\bar{\bH})}{(r-1)!}$ and $\bar{\bH}=\left[\begin{array}{ccc} \bH_{:,1}-\bH_{:,r}  & \ldots & \bH_{:,r-1}-\bH_{:,r}\end{array}\right]$.
In the noisy case, maximum likelihood optimization can be approximated as 
\begin{eqnarray}
\min_{\bH\in \mathbb{R}^{M \times r},\bS\in \Delta_r^N}\log(\text{vol}(\bH))+\frac{1}{\eta N}\|\bY-\bH\bS\|_F^2.
\end{eqnarray}
Therefore, the SSMF approach based on simplex volume minimization (SVmin) \cite{fu2016robust} can be posed as an approximation to the maximum likelihood problem. The same reference also introduces importance sampling and variational approaches to obtain the solution for the PRISM framework.

%polytope; and
%\item $\bV\in \mathbb{R}^{M \times N}$ is the noise matrix.
%\end{itemize}

%
% ,-.          .               ,--,       ;-.                            .    ,.                          .   
%(   `         |   o             /    o   |  )                           |   /  \                         |   
% `-.  ,-. ,-. |-  . ,-. ;-.    `.        |-'  ;-. ,-. ;-. ,-. ,-. ,-. ,-|   |--| ;-. ;-. ;-. ,-. ,-: ,-. |-. 
%.   ) |-' |   |   | | | | |      )   o   |    |   | | | | | | `-. |-' | |   |  | | | | | |   | | | | |   | | 
% `-'  `-' `-' `-' ' `-' ' '   `-'        '    '   `-' |-' `-' `-' `-' `-'   '  ' |-' |-' '   `-' `-` `-' ' ' 
%                                                      '                          '   '                       
%

\section{A Bayesian Perspective for Determinant Minimization Based SMF}
In this article, we propose a Bayesian formulation for the use of the determinant minimization criterion in structured matrix factorization frameworks such as SSMF \cite{fu2016robust} and PMF \cite{tatli:21tsp}. For this purpose, we assume that the left factor $\bH$ is a random matrix whose rows are drawn independently from a Gaussian distribution whose correlation matrix is posed as unknown. In fact, we model the correlation matrix as another random quantity. Section \ref{sec:datamodel} provides a description of the data model for the proposed Bayesian formulation. In Section \ref{sec:MAPestimation}, we show that the Maximumum A Posteriori (MAP) estimation of unknown variables in the data model is equivalent to a determinant minimization problem.
\subsection{Data Model}
\label{sec:datamodel}
The following underlies the data model for the proposed Bayesian framework: 
\begin{itemize}[leftmargin=*]
    \item We assume the generative model with the additive Gaussian noise model in (\ref{eq:SMFgenbaNoisy}),
    \item The left factor, 
     $\bH_g=\left[\begin{array}{cccc} \bh_1 & \bh_2 &  \ldots &  \bh_M \end{array}\right]^T$,
     is modeled as a random matrix whose (transpose of) row vectors $\bh_i \in \mathbb{R}^N$ for $i= 1$, \ldots, $M$ are independently drawn from the normal distribution $\mathcal{N}(\mathbf{0},\mathbf{\Sigma}_\bh)$, 
     
     \item The covariance matrix for the row vectors of the left factor matrix is also modeled as a random matrix with an inverse Wishart distribution, that is, $\mathbf{\Sigma}_\bh\sim \mathcal{IW}(\mathbf{\Psi},\varphi)$,  where
\begin{eqnarray*}
\hspace*{-0.04in}f(\mathbf{\Sigma}_\bh)=\frac{\det(\mathbf{\Sigma}_\bh)^{-\frac{r+\varphi+1}{2}}\exp{(-\frac{1}{2}\mathrm{Tr}\hspace{1pt}(\mathbf{\Psi}\mathbf{\Sigma}_\bh^{-1})})}{\det (\mathbf{\Psi})^{-\frac{\varphi}{2}}2^{\varphi\frac{r}{2}}\Gamma_r(\frac{\varphi}{2})},
\end{eqnarray*}
   which is the conjugate prior for Gaussian observations \cite{leonard1992bayesian,Gelman2013-jn}. Here, $\Gamma_r(\cdot)$ is the multivariate gamma function, and positive definite $\mathbf{\Psi}\in\mathbb{R}^{r \times r}$ and $\varphi\in\mathbb{R}$ are the scale matrix and the degrees of freedom parameters, respectively, for the inverse Wishart distribution. 
    
    \item $\{{\bS_g}_{:,j}, j= 1, \ldots, N \}$, the column vectors of the right factor, are drawn independently from the uniform distribution
    %\begin{eqnarray}
    %\label{eq:uniformS}
    $f_{\bs}(\bs)=\frac{1}{\text{vol}(\mathcal{D})}1_{\mathcal{D}}(\bs)$, 
%    \end{eqnarray}
    which are also independent of $\bH_g, \mathbf{\Sigma}_\bh$.

    \item The noise variables $\{{\bV}_{i,j}, i=1, \ldots, M, j=1, \ldots N\}$  are i.i.d. Gaussian random variables with zero mean and variance $\sigma_v^2$. Noise is assumed to be independent of all other variables in the model.
\end{itemize}

\subsection{Maximum A Posteriori Estimation Criterion}
\label{sec:MAPestimation}
Based on the data model presented in Section \ref{sec:datamodel}, we formulate the structured matrix factorization problem as the maximum a posteriori (MAP) estimation of variables $(\bH, \bS, \mathbf{\Sigma}_\bh)$:
\begin{eqnarray}
 \underset{\bH, \bS, \mathbf{\Sigma}_\bh}{\text{maximize}}& & \log f(\bH, \bS, \mathbf{\Sigma}_\bh|\bY), \label{eq:MAPfull}  
\end{eqnarray}
i.e., the maximization of the a posteriori pdf of unknowns $(\bH, \bS, \mathbf{\Sigma}_\bh)$ given the input $\bY$.
Using the Bayes rule, the MAP objective function, $\log f(\bH, \bS, \mathbf{\Sigma}_\bh|\bY)$, can be written more explicitly as
 \begin{equation*}
\begin{aligned}
& \log f(\bY|\bH,\bS,\mathbf{\Sigma}_\bh)+\sum_{k=1}^N\log f_\bs(\bS_{:,k})\nonumber \\&+\log f(\bH|\mathbf{\Sigma}_\bh)+\log f_{\mathbf{\Sigma}_\bh}(\mathbf{\Sigma}_\bh)-\log f_\bY(\bY).
\end{aligned}
\end{equation*}
%Now, we study each term separately. We note that 
%\begin{equation*}
%    \bY|\bH,\bS,\mathbf{\Sigma}_\bh \sim \bV,
%\end{equation*}
%since $\bY-\bH\bS = \bV$.
If we closely inspect the individual terms in the expression above:
\begin{itemize}[leftmargin=*]
\item Based on the observation model in (\ref{eq:SMFgenbaNoisy}), we can write 
\begin{equation*}
     \log f(\bY|\bH,\bS,\mathbf{\Sigma}_\bh)=-\frac{1}{2\mathbf{\sigma}_v^2}||\bY-\bH\bS||_F^2+c_1,
\end{equation*}
where $c_1=-MN\log(\sqrt{2\pi}\sigma_v)$. 

\item In the second term, $\log f_\bs(\bS_{:,k})$, acts as a barrier function to ensure that the ${\bS}_{:,k}$ vector is in $\mathcal{D}$. Therefore, the second term corresponds to the convex constraint 
%\begin{eqnarray}
 ${\bS}_{:,k} \in \mathcal{D}, \; k= 1, \ldots, N$.
%\end{eqnarray}

\item We can write the logarithm of the conditional distribution, $\log f(\bH|\mathbf{\Sigma}_\bh)$, more explicitly as
\begin{align}
 %  & -\frac{M}{2} \log \det(\mathbf{\Sigma}_\bh)-\frac{1}{2}\sum_{1}^{M}\bh_i^T\mathbf{\Sigma}_\bh^{-1}\bh_i+c_2, \nonumber \\
 %   &=-\frac{M}{2} \log \det(\mathbf{\Sigma}_\bh)-\frac{1}{2}\mathrm{Tr}\hspace{1pt}(\mathbf{\Sigma}_\bh^{-1}\sum_{1}^{M}\bh_i\bh_i^T)+c_2, \nonumber \\
    -\frac{M}{2} \log \det(\mathbf{\Sigma}_\bh)-\frac{1}{2}\mathrm{Tr}\hspace{1pt}(\mathbf{\Sigma}_\bh^{-1}\bH^T\bH)-\frac{Mr}{2}\log(2\pi). \label{eq:condfH}
\end{align}
%where $c_2=-\frac{Mr}{2}\log(2\pi)$. 

\item The logarithm of the Inverse-Wishart pdf, $\log f_{\mathbf{\Sigma}_\bh}(\mathbf{\Sigma}_\bh)$, can be written as
\begin{eqnarray}
&&\frac{\varphi}{2}\log \det (\mathbf{\Psi})-\frac{r+\varphi+1}{2}\log \det(\mathbf{\Sigma}_\bh)-\frac{r\varphi}{2}\log(2)\nonumber \\ &&-\log(\Gamma_r(\frac{\varphi}{2}))-\frac{1}{2}\mathrm{Tr}\hspace{1pt}(\mathbf{\Psi}\mathbf{\Sigma}_\bh^{-1}). \label{eq:logfSigmah}
\end{eqnarray}
\end{itemize}

We now examine the characterization of the MAP-optimal estimate of $\Sigma_\bh$. 
For this purpose, we derive the gradients of $\log f(\bH|\mathbf{\Sigma}_\bh)$ and $\log f_{\mathbf{\Sigma}_\bh}(\mathbf{\Sigma}_\bh)$ with respect to $\mathbf{\Sigma}_\bh$ and use the first-order optimality condition: 
%\begin{eqnarray*}
\begin{align*}
%&&\nabla(\log f(\bH|\mathbf{\Sigma}_\bh))_{\mathbf{\Sigma}_\bh^{-1}} = \frac{M}{2} \mathbf{\Sigma}_\bh-\frac{1}{2}\bH^T\bH, \label{bay1}\\
&\nabla_{\mathbf{\Sigma}_\bh}\log f(\bH|\mathbf{\Sigma}_\bh) = -\frac{M}{2} \mathbf{\Sigma}^{-1}_\bh+\frac{1}{2}\mathbf{\Sigma}_\bh^{-1}\bH^T\bH\mathbf{\Sigma}_\bh^{-1}, \\%\label{bay1}\\
&\nabla_{\mathbf{\Sigma}_\bh}\log f_{\mathbf{\Sigma}_\bh}(\mathbf{\Sigma}_\bh)= -\frac{r+\varphi+1}{2}\mathbf{\Sigma}_\bh^{-1}+\frac{1}{2}\mathbf{\Sigma}_\bh^{-1}\mathbf{\Psi}\mathbf{\Sigma}_\bh^{-1}. %\label{bay2}
\end{align*}
%&&\nabla(\log f(\mathbf{\Sigma}_\bh))_{\mathbf{\Sigma}_\bh^{-1}} = \frac{N+v+1}{2}\mathbf{\Sigma}_\bh-\frac{1}{2}\mathbf{\Psi}^T. \label{bay2}
%\end{eqnarray*}
Using the fact that the gradient of $\log f(\bH, \bS, \mathbf{\Sigma}_\bh|\bY)$ with respect to ${\mathbf{\Sigma}_{\bh}}$ at the optimal point $(\bH_*,\bS_*{\mathbf{\Sigma}_{\bh}}_*),$ is equal to zero, we obtain
\begin{eqnarray*}
\frac{M}{2} {\mathbf{\Sigma}_{\bh}}_*-\frac{1}{2}\bH_*^T\bH_* + \frac{r+\varphi+1}{2}{\mathbf{\Sigma}_{\bh}}_*-\frac{1}{2}\mathbf{\Psi} = \mathbf{0},  \label{bay3}
\end{eqnarray*}
From this expression, we obtain
\begin{eqnarray*}
    {\mathbf{\Sigma}_{\bh}}_*=\frac{1}{M+r+\varphi+1}(\bH_*^T\bH_*+\mathbf{\Psi}),
\end{eqnarray*}
as the relationship between the optimal values $\bH_*$ and ${\mathbf{\Sigma}_{\bh}}_*$.
Restricting ${\mathbf{\Sigma}_{\bh}}=\frac{1}{M+r+\varphi+1}(\bH^T\bH+\mathbf{\Psi})$,  the sum of $\log f(\bH|\mathbf{\Sigma}_\bh)$ in (\ref{eq:condfH}) and $\log f_{\mathbf{\Sigma}_\bh}(\mathbf{\Sigma}_\bh)$ in (\ref{eq:logfSigmah}) simplifies to 
 %we can reformulate MAP optimization in (\ref{eq:MAPfull}) only in terms of factors $(\bH,\bS)$ with the new objective function
%\begin{eqnarray*}
%\begin{aligned}
%&-\frac{M+r+\varphi+1}{2}\log \det(\frac{1}{M+r+\varphi+1}(\bH^T\bH+\mathbf{\Psi}^T)) \\
%&-\frac{1}{2}\mathrm{Tr}\hspace{1pt}(\mathbf{\Sigma}_\bh^{-1}(\bH^T\bH+\mathbf{\Psi}^T))+\log(\Gamma_r(\frac{\varphi}{2}))-\frac{r\varphi}{2}\log(2) \\ %&+\frac{\varphi}{2}\log \det (\mathbf{\Psi})-\frac{M+r+\varphi+1}{2}\log \det(\bH^T\bH+\mathbf{\Psi}).
%\end{aligned}
%\end{eqnarray*}
%If we eliminate the constant terms, this expression simplifies to
%\begin{eqnarray}
$-\frac{\beta}{2}\log \det(\frac{1}{\beta}(\bH^T\bH+\mathbf{\Psi}))+\text{constant terms}$,
%\end{eqnarray}
where $\beta=M+r+\varphi+1$ is a constant. As a result, ignoring the constant terms, the MAP-optimization in (\ref{eq:MAPfull}) reduces to the following determinant minimization problem:
%  \begin{equation*}
% \begin{aligned}
% & \underset{\bH, \bS, \mathbf{\Sigma}_\bh}{\text{maximize}}& & -\frac{1}{2\mathbf{\sigma}_v^2}||\bY-\bH\bS||_F^2-c_3\log \det(\bH^T\bH+\mathbf{\Psi}),  
% \end{aligned}
% \end{equation*}
\begin{mini!}[l]<b>
{\bH,\bS}
{\frac{1}{2\mathbf{\sigma}_v^2}||\bY-\bH\bS||_F^2+\frac{\beta}{2}\log \det(\frac{1}{\beta}(\bH^T\bH+\mathbf{\Psi}))\label{ff2}}{\label{ff}}{}
%\addConstraint{\bY=\bH \bS}{\label{eq:detminpconstr1}}{}
\addConstraint{\bS_{:,j} \in \mathcal{D}}{,\quad}{j= 1, \ldots, N.\label{eq:detminplconstr2}}{ }
\end{mini!}
The optimization problem in (\ref{ff}) is in the same form as the robust simplex volume minimization problem in \cite{fu2016robust} for the domain $\mathcal{D}=\Delta_r$ and for the choice $\mathbf{\Psi}=\tau \mathbf{I}$. This choice is motivated from an algorithmic advantage point of view and is introduced as a computational heuristic. 

Similarly, the determinant minimization problem in (\ref{ff}) is also proposed for the PMF framework for identifiable polytopes as domains \cite{tatli:21tsp,tatli:21icassp} and for the choice $\mathbf{\Psi}=\tau \mathbf{I}$.

\section{Discussion of Results}

The Bayesian framework introduced in this article provides a probabilistic justification for the geometrically introduced optimization in (\ref{ff}) proposed for structured matrix factorization problems\cite{fu2016robust, tatli:21tsp}. We can make the following observations on the basis of this result:
\begin{itemize}[leftmargin=*]
    \item \looseness=-1 If we define $\hat{\mathbf{\Sigma}}_\bh^{(s)}=\frac{1}{M}\sum_{i=1}^{M}\bh_i\bh_i^T$ as the sample-based estimate of the covariance of the rows of $\bH$, then the argument of the $\log\det(\cdot)$ function in (\ref{ff2}) can be rewritten as
  %  \begin{eqnarray*}
    \begin{align*}
    \frac{1}{\beta}(\bH^T\bH+\mathbf{\Psi})&=\frac{1}{\beta}(M\hat{\mathbf{\Sigma}}^{(s)}_\bh+(\varphi+r+1)\frac{\mathbf{\Psi}}{\varphi+r+1}),\\
    &=\underbrace{\mu \hat{\mathbf{\Sigma}}^{(s)}_\bh+(1-\mu) \hat{\mathbf{\Sigma}}_\bh^{\text{mode,prior}}}_{\hat{\mathbf{\Sigma}}_\bh},
    \end{align*}
    where $\mu=\frac{M}{M+\varphi+r+1}$. Therefore, it is equivalent to the convex combination of the sample covariance $\hat{\mathbf{\Sigma}}_\bh^{(s)}$ and the mode of prior pdf for covariance, $\hat{\mathbf{\Sigma}}_\bh^{\text{mode,prior}}=\frac{\mathbf{\Psi}}{\varphi+r+1}$, which is the MAP estimate of the covariance based only on the prior distribution. We can rewrite the scaled version of the objective in (\ref{ff2}), ignoring the constant terms, as
    \begin{eqnarray}
     J(\bH,\bS)=\|\bY-\bH\bS|_F^2+\lambda \log\det(\hat{\mathbf{\Sigma}}_\bh) \label{eq:ovobject},
    \end{eqnarray}
    where $\lambda=(M+r+\varphi+1)\sigma_v^2$, which is a $\log\det$-barrier-regularized least squares problem. The argument of the $\log\det$ function is essentially a covariance estimate for the row vectors of $\bH$, where the estimate would lean more towards the sample estimate $\hat{\mathbf{\Sigma}}_\bh^{(s)}$ for higher input dimensions.\looseness=-1
 %   \end{eqnarray*}
 \item The proposed framework provides a prescription for the choice of the parameter $\lambda$ of the $\log\det$ regularizer in (\ref{eq:ovobject}) by showing its dependence on the noise variance and the dimensions of the input and latent vectors. 
 %Furthermore, we obtain an analytical formulation for the scaling of the $\log \det$ regularizer term added to the least squares reconstruction term in (\ref{eq:ovobject}) in terms of the noise level and the dimensions of the input and latent vectors.
    
\end{itemize}

\section{Numerical Experiment}
To validate the analytical results obtained from the proposed framework, we performed an experiment with $\mathcal{D}=\mathcal{B}_\infty=\{\bx \hspace{0.02cm}\vert\hspace{0.02cm} \|\bx\|_\infty\le 1, \bx\in\mathbb{R}^r\}$, latent vector dimension $r=5$, input dimension $M=20$, the number of input vectors $N=1000$ and noise level $\sigma_v^2=0.01$. We generate the rows of $\bH_g$ from $\mathcal{N}(\mathbf{0},\mathbf{I})$. For the prior density scaling matrix, we consider $\mathbf{\Psi}=\rho (\varphi+r+1)\mathbf{I}$, where $\rho$ is the algorithm hyperparameter. Note that, only for $\rho=1$, the prior-based MAP estimate $\hat{\mathbf{\Sigma}}_h^{mode,prior}$ would match the data generation model. We evaluate the performance of the MAP estimate based on (\ref{eq:ovobject}) for different hyperparameter choices: $\varphi\in\{6,250\}$ and $\rho \in [10^{-4},10^{2}]$. We choose $\lambda=(M+r+\varphi+1)\sigma_v^2$ as suggested by the framework. We use the coordinate descent, in the parameters $\bH$ and $\bS$, for the objective in (\ref{eq:ovobject}), using the Nesterov accelerated gradient method.

\begin{figure}[h]
\begin{center}
% trim={<left> <lower> <right> <upper>}
    \includegraphics[width=7cm]{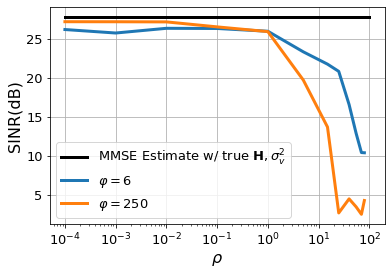}
    \caption{SINR performance for estimating $\bS_g$.}
    \label{fig:simresults}
\end{center}
\end{figure}
Figure \ref{fig:simresults} shows the signal-to-interference-plus-noise-power-ratio (SINR) performances in estimating $\bS_g$ as a function of $\rho$ for two different choices of $\varphi$. The same figure shows the SINR of the best MMSE estimator with perfect knowledge of $\bH_g$ and $\sigma_v^2$ as a utopia-benchmark. From this figure we observe that the performance decreases significantly when the $\mathbf{\Psi}$ parameter is selected higher than the nominal value ( $\rho>1$) while we observe almost no performance drop for $\rho<1$. The results of the experiment also confirm the validity of the regularization constant ($\lambda$) prescription of the framework. % Furthermore, for higher $\varphi$ choices, the performance is better (compared to smaller $\varphi$ choices), when $\rho<1$ and worse when $\rho>1$.

\vfill\pagebreak

% References should be produced using the bibtex program from suitable
% BiBTeX files (here: strings, refs, manuals). The IEEEbib.bst bibliography
% style file from IEEE produces unsorted bibliography list.
% -------------------------------------------------------------------------
\bibliographystyle{IEEEbib}
\bibliography{article}

\end{document}